\documentclass[preprint,review,12pt]{elsarticle}
\usepackage{amssymb}
\usepackage{amsmath}
\usepackage{array}
\usepackage{booktabs}
\usepackage{xcolor}

\usepackage{graphicx} 
\usepackage{multirow}
\usepackage{fullpage}
\usepackage{tabularx}
\usepackage{rotating, makecell}
\usepackage{svg}
\usepackage{algpseudocode}

\begin{document}
\begin{frontmatter}

\title {An Embedded Intelligent System for Attendance Monitoring} 
 \author{TOUZENE Abderraouf}
\author{ABED Abdeljalil Wassim}
\author{Slimane LARABI}
\ead{slarabi@usthb.dz}
\affiliation{organization={Computer Science Faculty, USTHB University}, addressline={BP 32}, 
            city={El Alia},
            postcode={16111}, 
            state={Algiers},
            country={Algeria}}
\begin{abstract}
In this paper, we propose an intelligent embedded system for monitoring class attendance and sending the attendance list to a remote computer. The proposed system consists of two parts : an embedded device (Raspberry with PI camera) for facial recognition and a web application for attendance management. The proposed solution take into account the different challenges: the limited resources of the Raspberry Pi, the need to adapt the facial recognition model and achieving acceptable performance using images provided by the Raspberry Pi camera.
\end{abstract}


\end{frontmatter}

\section{Introduction}

Managing attendance in educational institutions represents a major logistical challenge. Traditionally, manual attendance taking is not only time-consuming but also prone to human errors and potential fraud. To overcome these limitations, facial recognition has emerged as a promising technology to automate this process while ensuring increased reliability.
However, implementing facial recognition systems on embedded devices like the Raspberry Pi presents significant challenges. The Raspberry Pi, with its limited computational power and memory resources, requires specific optimizations to efficiently run deep learning models that are typically resource-intensive. Furthermore, deployment in real-world environments, such as classrooms, imposes additional constraints such as variability in lighting conditions and the need for processing.\\

In this paper we propose an embedded system for attendance monitoring based on facial recognition. 
In the past, computer vision applications were primarily executed on traditional computers \cite{dahmane2012, setitra2015, larabi2018, zatout2022}. However, with the increase in computational power of devices, edge computing has become very useful, enabling enhanced data security. \\
The proposed system is designed to capture images of participants, process these images to detect and recognize faces, and record attendance data via a web application. The primary goal is to provide a practical and reliable solution for student identification in the classroom while addressing the specific constraints of resource-limited environments.\\

It aims to demonstrate the feasibility and effectiveness of such an approach in an educational setting, providing a basis for future improvements and potential extensions.\\
This paper is organized as follows: section 2 presents some methodologies used for facial detection and representation, discussing classical approaches as well as modern deep learning-based models. Section 3 present the stages implemented in this work. In section 4 experimental results obtained are presented are discussed. Finally section 5 conclude the paper with some remarks.

\section{Related Works}
\subsection{Face Detection Methods}

Facial detection is the process of determining whether a face is present in an image. Several facial detection methods can be mentioned:

\noindent\textbf{- Knowledge-based face detection:}\\
These methods rely on a set of rules developed by humans. A face should have a nose, eyes, and a mouth at certain distances and positions relative to each other. The problem with these methods lie in constructing an appropriate set of rules. If the rules are too general or too detailed, the system ends up with many false positives. Furthermore, these methods do not work for all skin colors and depend on lighting conditions, which can alter the exact hue of the skin in the image.

\noindent\textbf{- Template matching:}\\
Template matching uses predefined or parameterized face models to locate or detect faces by correlating the predefined or deformable models with the input images.

\noindent\textbf{- Feature-based face detection:}\\
This kind of methods extracts structural features of the face. It is trained as a classifier and then used to differentiate facial and non-facial regions. An example of these methods is color-based face detection, which analyzes colored images or videos to spot areas with typical skin color and then searches for face segments. Among the most well-known approaches using this method are Haar cascades and Histograms of Oriented Gradients (HOG).

\noindent\textbf{- Appearance-based face detection:}\\
This kind of methods uses a set of training face images to find face models. It relies on machine learning and statistical analysis to find relevant features of face images and extract characteristics. These methods use several algorithms such as PCA, Hidden Markov Models, Naive Bayesian Classification, and neural networks.

The evolution of face detectors has been marked by a shift from approaches based on classical algorithms to those based on deep learning, leading to significant advances in terms of accuracy, speed, and robustness in detecting faces under varied conditions. Classical algorithms like Haar cascades \cite{haar} and HOG \cite{hog} constituted the initial breakthroughs in face detection. However, these approaches had limitations, notably less satisfactory performance for noisy images and varying lighting conditions.

To overcome these challenges, deep learning-based detectors have emerged as more powerful and versatile solutions. Models such as SSD \cite{ssd}, MTCNN \cite{mtcnn}, RetinaFace \cite{retinaface}, MediaPipe \cite{mediapipe}, YuNet \cite{yunet}, and YoloV8 \cite{yolov8} have revolutionized the field by exploiting the capabilities of deep neural networks to learn highly discriminative features from massive training data. For example, SSD (Single Shot Multibox Detector) introduced an innovative approach that allows detecting objects in a single pass over an input image, significantly improving detection efficiency. Similarly, MTCNN (Multi-Task Cascaded Convolutional Neural Network) proposed a cascaded architecture of interconnected neural networks enabling precise face detection and facial landmark identification.

\subsection{Representation}

The models used for facial representation are classical convolutional neural networks (CNNs). They represent facial images in the form of vectors \cite{deepface}.

Among the most efficient models are DeepFace \cite{taigman2014deepface} by Facebook, launched in 2014, which was one of the first to adopt deep learning techniques for an in-depth facial analysis, achieving an impressive prediction accuracy of 97.35\%. Followed closely by FaceNet \cite{facenet} by Google in 2015, which introduced a compact face representation in Euclidean space, thus facilitating clustering and recognition with accuracy reaching 99.25\% on certain test datasets. Oxford University also made its mark with VGG-Face \cite{vgg}, a robust model using Triplet loss function to learn face embeddings. With a prediction accuracy reaching 98.95\% on LFW (Labeled Faces in the Wild) \cite{Huang2012a}, it cemented its place among leading models. Concurrently, ArcFace \cite{deng2018arcface} introduced an innovative architecture with a dedicated fully connected layer, further increasing prediction accuracy to 99.40\% on the LFW dataset.

More recently, SFace \cite{sface}, introduced in 2021, made a significant contribution by proposing a new loss function, the hypersphere constraint loss, which outperformed its predecessors on various public datasets with an impressive execution speed. These models have not only redefined facial recognition standards but also opened new opportunities in fields such as security, surveillance, and computer vision, offering promising prospects for the future of this technology.

\section{Proposed System}

\begin{figure}[ht]
    \centering
    \includegraphics[width=14cm]{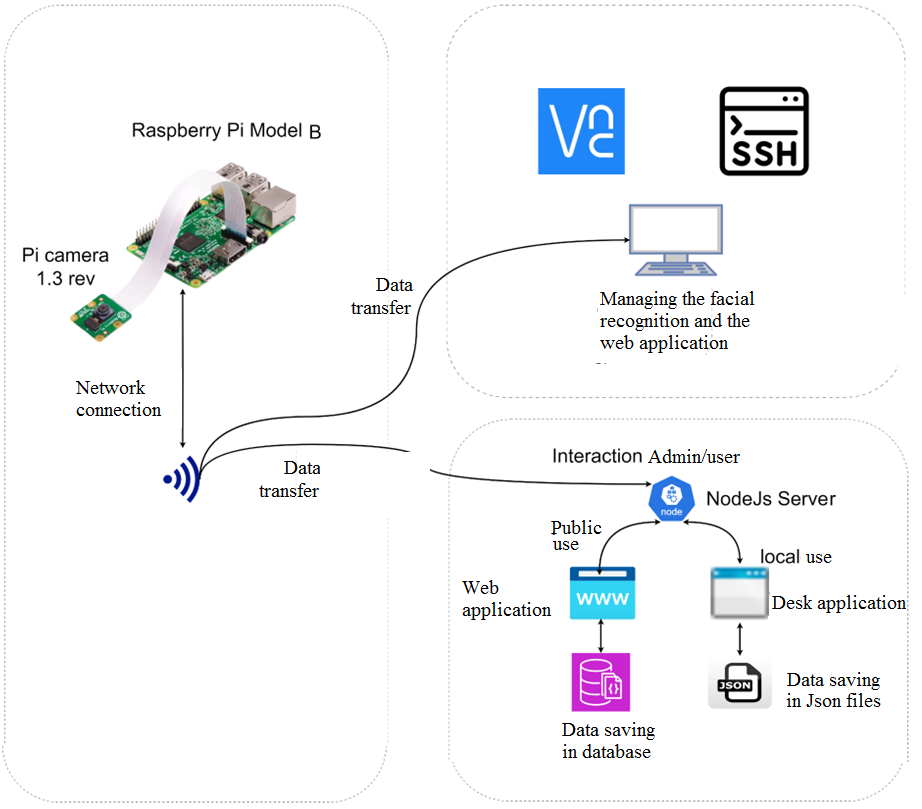}
    \caption{Overall system architecture}
    \label{fig:archisys}
\end{figure}

The system architecture includes a facial recognition system, which operate in network mode, thereby enabling more reliable transfer and management of attendance data. This data is transferred to a database hosted on a server connected to the same network. \\
In the absence of a network or server, the system will function in hotspot mode, where it will create its own local network, allowing direct connection with the desktop application on the user’s computer. \\
Developers can interact with the system via SSH (Secure Shell) and VNC (Virtual Network Computing) protocols for maintenance and deployment. Administrators and end users can access the system through a user-friendly interface, either via a web application or a desktop application, depending on the chosen connection method. Figure \ref{fig:archisys} illustrates a schematic representation of this architecture.

\subsection{Reference Image Storage}
To add a student's image to our database, we first apply a sharpness filter to enhance the image quality. This step improves the capture and representation of relevant facial features. Typically, the images used come from identity photos or student badges, which are often low in detail and slightly blurred. \\
Next, we use a facial representation model to extract a feature vector from the image. Once this encoding is obtained, it will be stored in our database, where it will serve as a reference for comparison with other images for student identification. 

\subsection{Facial Recognition System}
The embedded facial recognition system we have developed follows the following steps:\\

The system starts by capturing a video using the Pi Camera rev 1.3 module of the Raspberry Pi, which will be processed in the subsequent steps of our system.

The first step of our system is to extract the coordinates of the faces as well as the key points from the video. It outputs the coordinates of the bounding boxes and the key points for each detected face.

An object tracking algorithm has been implemented in our system to address two essential objectives. Firstly, to reduce the number of requests transmitted to the attendance monitoring system: by tracking the movements of detected faces, we can generate one request per face rather than sending a request for each new detection, thereby lightening the system's load. \\
Secondly, the tracking algorithm also helps minimize the number of false positives. Instead of simply detecting a face and automatically assigning an identification to each new appearance, the system keeps track of identifications already made for a given face, thus avoiding the generation of repeated false identifications for the same person at different times. \\
This approach, therefore, contributes to strengthening the system's accuracy while ensuring more efficient data and resource management. \\
To meet our performance constraints, we have opted to use the centroid-based object tracking algorithm \cite{tracker}. 

\begin{figure}[ht]
    \centering
    \includegraphics[width=10cm]{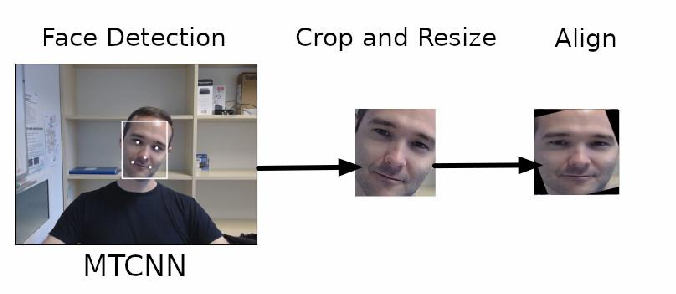}
    \caption{Example of alignment and cropping of a face \cite{alignement}}
    \label{fig:alignment}
\end{figure}

Alignment and Cropping:\\
In this step, we calculate and then apply a geometric transformation (scaling, rotation, and translation) that aligns the facial landmarks (extracted from the detection step) of the source image to a standardized position. It produces a new image as output in which the facial landmarks are aligned and cropped, with the aim of increasing the accuracy of facial recognition. Researchers at Google have stated that the application of an alignment algorithm increased the accuracy of their facial recognition model from 98.87\% to 99.63\% \cite{deepface}. Figure \ref{fig:alignment} illustrates an example of alignment.

\begin{figure}[ht]
    \centering
    \includegraphics[width=0.8\textwidth]{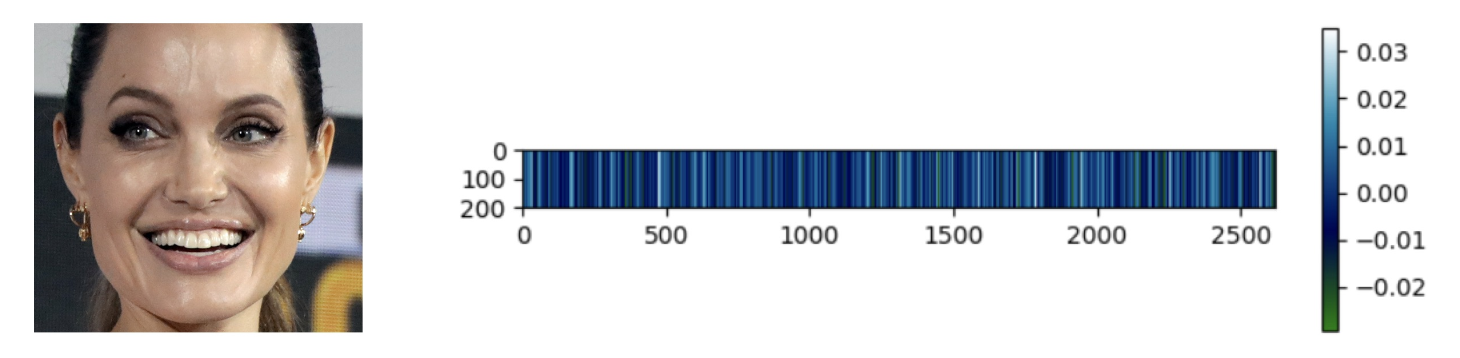}
    \caption{Example of face representation \cite{deepface}}
    \label{fig:representation}
\end{figure}

Representation:\\
A deep learning model is used to extract facial features to obtain a vector representation of the face. Figure \ref{fig:representation} shows an example of a vector representation of a face.

Computing Distances:\\
After the representation step, we proceed to calculate the distances between the extracted facial features and the features of faces stored in the database. The minimum distance is then compared to a pre-defined threshold experimentally to determine whether the face is recognized or not.

\section{Experimental results}
To deploy the described system on a Raspberry Pi, we need to consider the constraints imposed by the limited resources of this embedded system. Therefore, we followed the following steps:

\subsection{Implementation details}

- We first adopted an approach based on the use of threads to effectively separate video input management from the facial recognition process. This approach allowed optimal allocation of available resources, thus ensuring smooth processing of video streams without any frame loss despite hardware constraints.

- In order to minimize processing load while maintaining acceptable inference times and without compromising the system's accuracy, we decided to reduce the resolution of captured videos.

- ONNX Runtime is designed to provide high performance for executing deep learning models on various hardware platforms. By using techniques such as graph optimization, model quantization, and efficient thread management, ONNX Runtime aims to reduce inference time and improve computational efficiency. These optimizations are particularly important for deployments on devices like the Raspberry Pi, where resources are limited, and performance needs to be maximized.

- To make the use of our system simpler, dedicated services have been set up on the operating system to manage:

\begin{itemize}
    \item \textbf{Network connection Startup:} Connection startup can be done either via a local network, Wi-Fi, or a wired connection (Ethernet), where the Raspberry Pi is connected to other devices through the network, or in hotspot mode where it is directly connected with the host machine via Wi-Fi.

    \item \textbf{Startup of the Facial Recognition System:} A service has been created to ensure the automatic launch of the presence system scripts during the startup of the embedded system.

    \item \textbf{Startup of the Configuration Interface:} We have set up a configuration interface allowing users to adjust system parameters according to their specific needs to meet all scenarios, as well as configure the connection and ensure data transfer from the embedded system to the presence management system.
\end{itemize}

\subsection{Attendance Management}
For classroom attendance management, we developed two complementary applications using the MERN stack (MongoDB, Express.js, React, Node.js) for the web application and a file-based system for the desktop application.\\

The web application, built with React, offers a responsive and modern user interface for online attendance management. Teachers and administrators can view real-time attendance, generate reports, and manage student information. Data is centralized in a MongoDB database, providing flexible and efficient document management. The backend of the application, developed with Express.js and Node.js, handles user requests, authentication, and communication with the database, ensuring the security and integrity of information. This setup allows for real-time management and a smooth, secure user experience.\\

The desktop application, developed using Electron, is designed to work independently without requiring a constant internet connection. This application uses React for a consistent user interface with the web application. Unlike the web application, it stores data locally as files, allowing for local processing and data management without relying on a centralized database. Users can import attendance data captured by the facial recognition system, analyze this data locally, and export it for later use or synchronization with the web application. This approach is particularly useful for environments with limited connectivity, ensuring that teachers can manage and analyze attendance flexibly.\\

By integrating these two applications, we provide a comprehensive solution for classroom attendance management, combining the benefits of centralized online management with the flexibility of local processing depending on network availability.

\subsection{Conducted experiments and Obtained results}

\begin{figure}[ht]
    \centering
    \includegraphics[width=14cm]{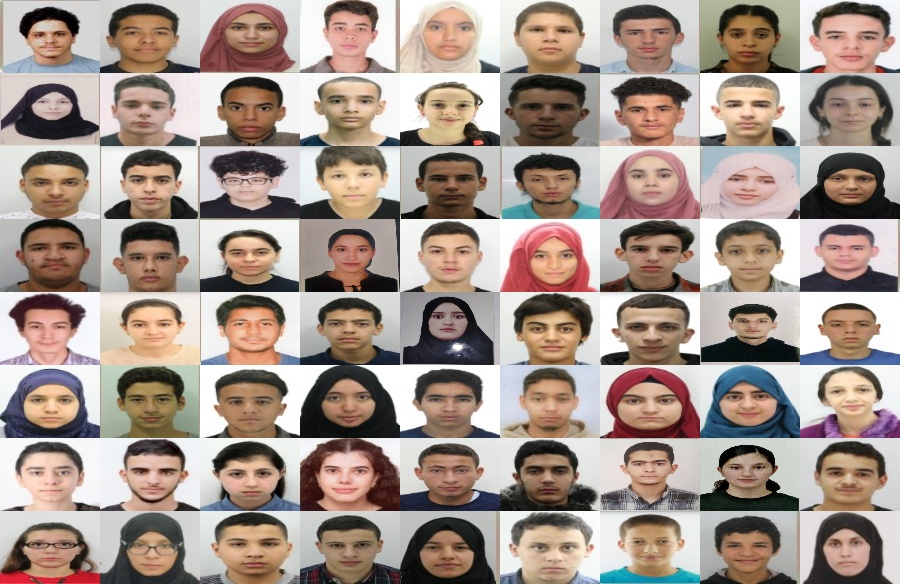}
    \caption{Some images from used data.}
    \label{fig:dataset}
\end{figure}

The dataset we used to test the performance of our system consists of 77 images. These images are photos faculty student. Figure \ref{fig:dataset} illustrates some images from our dataset.

For the choice of detection and recognition models, we compared and tested several models. Initially, we chose the DeepFace module \cite{deepface} which we evaluated using different configurations, particularly on two sets of data (77 and 7 student images). We conducted tests on five separate videos to assess the model's performance in different scenarios:

\begin{description}
    \item[Scenario 1:] features a student without any accessory covering part of their face.
    \item[Scenario 2:] features a single student wearing a cap.
    \item[Scenario 3:] features a single student wearing a hoodie.
    \item[Scenario 4:] features a single student whose facial feature vector is not in the database.
\end{description}

Tables \ref{tab:deepface7} and \ref{tab:deepface70} present a comparison of the performance of some models implemented by DeepFace \cite{deepface} across all the aforementioned scenarios.

\begin{table}[ht]
    \centering
    \caption{Comparison of the performance of models implemented by DeepFace with 7 student images.}
    \label{tab:deepface7}
    \begin{tabular}{|c|c|c|c|c|c|c|}
        \hline
        \multicolumn{2}{|l|}{{Recognition}{------Detection}} &  Arc Face \cite{deng2018arcface} & Facenet 512 & Facenet \cite{facenet}& Dlib & VGG-Face \cite{vgg} \\
        \hline
        yolov8 \cite{yolov8} & cosine & 82.91\% & 52.99\% & 80.34\% & 11.97\% & 82.05\%\\
        & euclidean & 82.05\% & 82.91\% & 82.05\% & 70.09\% & 82.05\% \\
        \hline
        yunet \cite{yunet} & cosine & 5.13\% & 3.42\% & 2.56\% & 5.13\% & 5.13\%\\
        & euclidean & 2.56\% & 5.98\% & 80.34\% & 5.13\% & 5.13\% \\
        \hline
        Fast MTCNN & cosine & 0.00\% & 65.81\% & 76.92\% & 72.41\% & 82.05\% \\
        & euclidean & 0.00\% & 82.05\% & 80.34\% & 75.21\% & 82.05\% \\
        \hline
        Dlib & cosine & 82.05\% & 85\% & 76.92\% & 3.42\% & 35.90\% \\
        & euclidean & 48.72\% & 5.98\% & 80.34\% & 5.13\% & 49.57\% \\
        \hline
        SSD \cite{ssd} & cosine & 1.71\% & 85\% & 0.00\% & 0.00\% & 85\% \\
        & euclidean & 0.00\% & 1.71\% & 0.00\% & 0.00\% & 0.85\% \\
        \hline
        MTCNN \cite{mtcnn} & cosine & 82.05\% & 85\% & 9.40\% & 11.11\% & 17.09\%\\
        & euclidean & 19.66\% & 8.55\% & 81.20\% & 14.53\% & 17.09\% \\
        \hline
        Mediapipe \cite{mediapipe} & cosine & 17.95\% & 1.71\% & 20.51\% & 17.95\% & 18.80\% \\
        & euclidean & 16.24\% & 80.34\% & 22.22\% & 17.95\% & 18.80\% \\
        \hline
    \end{tabular}
\end{table}

\begin{table}[ht]
    \centering
    \caption{Comparison of the performance of models implemented by DeepFace with 77 student images.}
    \label{tab:deepface70}
    \begin{tabular}{|c|c|c|c|c|c|c|}
        \hline
        \multicolumn{2}{|l|}{{Recognition}{-------Detection}} & Arc Face \cite{deng2018arcface}& Facenet 512 & Facenet \cite{facenet} & Dlib & VGG-Face \cite{vgg} \\
        \hline
        yolov8 \cite{yolov8} & cosine & 5,13\% & 0,00\% & 0,00\% & 2,56\% & 1,71\% \\
        & euclidean & 1,71\% & 85\% & 0,00\% & 2,56\% & 1,71\% \\
        \hline
        yunet \cite{yunet} & cosine & 4,27\% & 0,00\% & 0,85\% & 4,27\% & 1,71\% \\
        & euclidean & 85\% & 85\% & 0,00\% & 5,13\% & 1,71\% \\
        \hline
        Fast MTCNN & cosine & 4,31\% & 0,00\% & 0,00\% & 3,45\% & 1,72\% \\
        & euclidean & 2,59\% & 1,71\% & 0,00\% & 10,34\% & 1,72\% \\
        \hline
        Dlib & cosine & 85\% & 0,00\% & 0,00\% & 0,00\% & 2,56\% \\
        & euclidean & 85\% & 85\% & 0,00\% & 1,71\% & 2,56\% \\
        \hline
        SSD \cite{ssd}& cosine & 1,71\% & 0,00\% & 0,00\% & 0,00\% & 0,85\% \\
        & euclidean & 0,00\% & 85\% & 0,00\% & 0,00\% & 0,85\% \\
        \hline
        MTCNN \cite{mtcnn}& cosine & 2,56\% & 0,00\% & 0,00\% & 4,27\% & 1,71\% \\
        & euclidean & 1,71\% & 85\% & 0,00\% & 4,27\% & 1,71\% \\
        \hline
        Mediapipe \cite{mediapipe}& cosine & 17,09\% & 0,85\% & 17,09\% & 17,09\% & 17,09\% \\
        & euclidean & 16,24\% & 17,09\% & 17,09\% & 17,09\% & 17,09\% \\
        \hline
    \end{tabular}
\end{table}

Increasing the number of vectors clearly impacted the performance of the models. Therefore, we decided to try other models, namely Yunet \cite{yunet} and SFace \cite{sface}, since increasing the number of students from 7 to 77 did not impact the recognition rate as shown in Table \ref{tab:sface}.

\begin{table}[ht]
    \centering
    \begin{tabular}{|c|c|c|c|c|c|c|c|c|}
      \hline
    Scenarios & \begin{tabular}[c]{@{}c@{}}Correct\\ Recognition\end{tabular} & \begin{tabular}[c]{@{}c@{}}Incorrect\\ Recognition\end{tabular} & \begin{tabular}[c]{@{}c@{}}Unknown\end{tabular} & \begin{tabular}[c]{@{}c@{}}Detected\\ Faces\end{tabular} & ACC & FAR & FRR \\ \hline
      Scenario 1 & 14 & 0 & 3 & 17  & 82\% & 0\% & 0.18\%\\ \hline
      Scenario 2 & 3 & 0 & 9 & 12 & 25\% & 0\% & 75\%\\ \hline
      Scenario 3 & 2 & 0 & 10 & 12 & 17\% & 0\% & 83\%\\ \hline
      Scenario 4 & 0 & 0 & 24 & 24 & 100\% & 0\% & 0\% \\ \hline
    \end{tabular}
    \caption{Results of tests of the Yunet and SFace models with 77 and 7 student images where  (ACC):accuracy, (FAR): false acceptance rate , and (FRR): false rejection rate.}
    \label{tab:sface}
\end{table}

\subsection{Robustness of the system to Camera pose and Light conditions}

We began by building a database comprising the facial characteristic vectors of 77 students, using their identity photos. These characteristic vectors were used as references for the identification. The experiments took place in two different practical work rooms with two groups of 6 and 9 students participated in this experiment. We conducted tests on three scenarios to evaluate the system's performance under real conditions.

\begin{enumerate}
    \item \textbf{First Case: Optimal Viewing Angle with Poor Lighting Conditions} \\
In this first experimental case, the lighting conditions were poor, resulting in dark faces of the subjects in the captured images. Despite these difficulties, the system detected the five entering students, demonstrating its robustness and ability to function effectively even in light conditions. Figure \ref{fig:scene1l} shows the images of the recognized faces and the scene used for this scenario.

\begin{figure}[ht]
\centering
     \includegraphics[width=10cm]{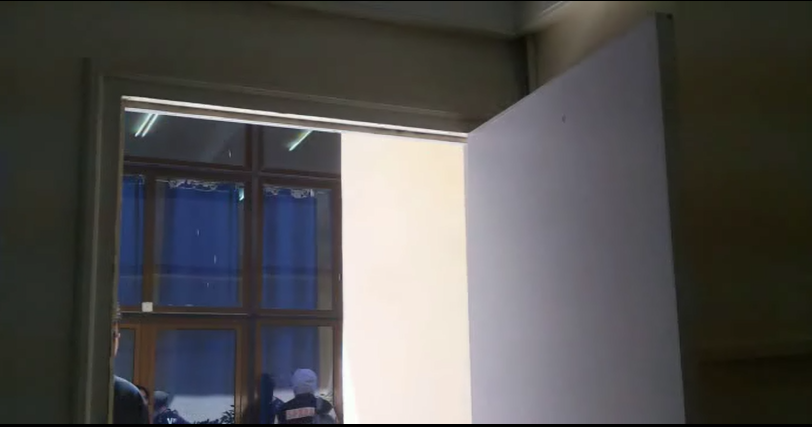}\\
     \includegraphics[width=10cm]{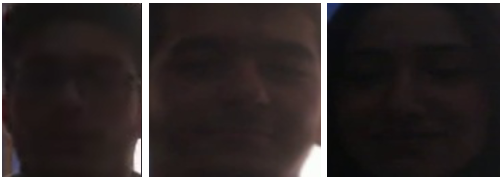}
     \includegraphics[width=8cm]{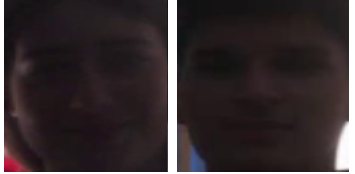}
       \caption{Scene of the first scenario and obtained results}
        \label{fig:scene1l}
\end{figure}

    \item \textbf{Second Case: Optimal Viewing Angle with Acceptable Lighting Conditions} \\ 
In this second case, the shooting angle was optimal, and the lighting conditions were acceptable. Under these favorable conditions, the system achieved accurate detection and successfully identified the six present individuals. Figure \ref{fig:scene2l} shows the images of the recognized faces and the scene used for this scenario.

\begin{figure}[ht]
\centering
        \includegraphics[width=10cm]{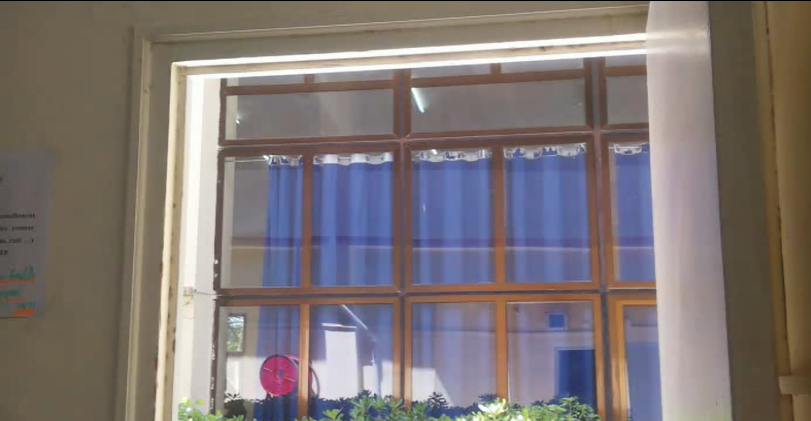}
        \includegraphics[width=10cm]{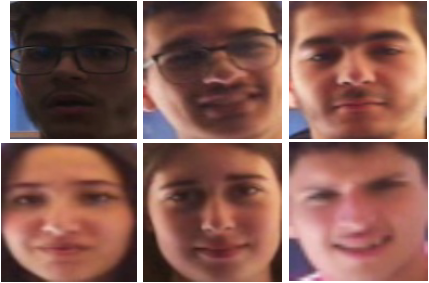}
        \label{fig:scene2l}
   \caption{The second case and obtained results.}
\end{figure}

\item \textbf{Third Case: Unfavorable Viewing Angle with Acceptable Lighting Conditions} \\ 
In this third case, although the lighting conditions were better than in the first case, the viewing angle was not optimal, making detection more difficult for the system. Despite this, the system was able to detect 6 from the 9 entering students, highlighting the importance of the viewing angle in the recognition process. Figure \ref{fig:scene3l} shows the images of the recognized faces, the images of unrecognized faces, and the scene used for this scenario.

\begin{figure}[ht]
\centering
    \includegraphics[width=10cm]{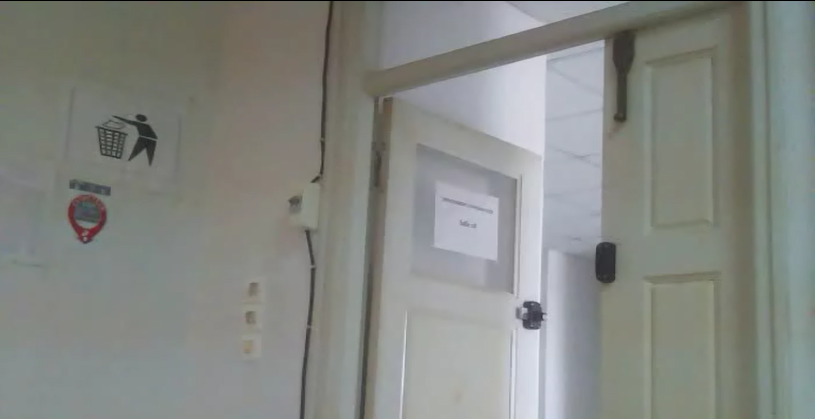}\\
    \includegraphics[width=10cm]{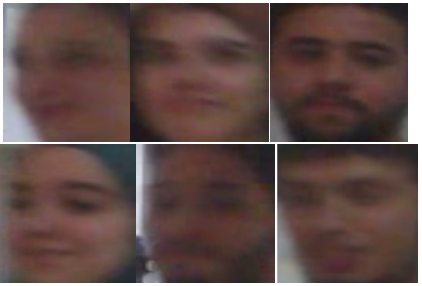}\\
    \includegraphics[width=10cm]{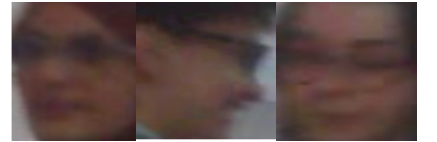}
    \caption{Third scene and obtained results}
    \label{fig:scene3l}
\end{figure}

\end{enumerate}
\section*{Conclusion}

This we presented a facial recognition-based attendance monitoring system.

Initially, we focused our efforts on deploying and optimizing pretrained models on the embedded system, emphasizing their efficient integration and adaptation to specific hardware constraints. This enabled us to achieve satisfactory facial recognition performance under limited resource conditions.

Subsequently, we designed and developed a user-friendly web application for managing the attendance system. The interaction between the embedded system and the web application facilitated real-time attendance management, making data access and system administration easier.

We also developed a desktop application to provide an alternative in case of unavailability of a computer network.

future work will focus on using more advanced optimization techniques to optimize resource management by the operating system and improving the quality of images acquired by the system, either by using a more powerful camera module or by developing an optimized super-resolution model suitable for embedded systems. These enhancements would contribute to the overall robustness and efficiency of the attendance monitoring system, making it more reliable and effective in various scenarios.


\begin{thebibliography}{10}

\bibitem{dahmane2012}
A.~Dahmane, S.~Larabi, C.~Djeraba, and I.~M. Bilasco, ``Learning symmetrical model for head pose estimation.,'' in {\em ICPR - 21st International Conference on Pattern Recognition, Tsukuba, Japan}, pp.~3614--3617, 2012.

\bibitem{setitra2015}
I.~Setitra and Larabi., ``Sift descriptor for binary shape discrimination, classification and matching.,'' in {\em International Conference on Computer Analysis of Images and Patterns}, pp.~489--500, 2015.

\bibitem{larabi2018}
S.~Larabi, ``Augmented reality for mobile devices: Textual annotation of outdoor locations.,'' in {\em Augmented Reality and Virtual Reality: Empowering Human, Place and Business.}, pp.~353--362, 2018.

\bibitem{zatout2022}
C.~Zatout and S.~Larabi, ``Semantic scene synthesis: application to assistive systems,'' {\em The Visual Computer 38 (8)}, pp.~2691--2705, 2022.

\bibitem{haar}
A.~Sharifara, M.~S. Mohd~Rahim, and Y.~Anisi, ``A general review of human face detection including a study of neural networks and haar feature-based cascade classifier in face detection,'' in {\em 2014 International Symposium on Biometrics and Security Technologies (ISBAST)}, (Kuala Lumpur, Malaysia), pp.~73--78, 2014.

\bibitem{hog}
S.~G. Joshi and R.~Vig, ``Histograms of orientation gradient investigation for static hand gestures,'' in {\em International Conference on Computing, Communication \& Automation}, (Greater Noida, India), pp.~1100--1103, 2015.

\bibitem{ssd}
W.~Liu, D.~Anguelov, D.~Erhan, C.~Szegedy, S.~Reed, C.-Y. Fu, and A.~C. Berg, ``Ssd: Single shot multibox detector,'' {\em UNC Chapel Hill}, 2016.

\bibitem{mtcnn}
K.~Zhang, Z.~Zhang, Z.~Li, and Y.~Qiao, ``Mtcnn: Face detection alignment,'' {\em IEEE Signal Processing Letters}, 2016.

\bibitem{retinaface}
J.~Deng, J.~Guo, Y.~Zhou, J.~Yu, I.~Kotsia, and S.~Zafeiriou, ``Retinaface: Single-stage dense face localisation in the wild,'' {\em Imperial College London}, 2019.

\bibitem{mediapipe}
C.~Lugaresi, J.~Tang, H.~Nash, C.~McClanahan, E.~Uboweja, M.~Hays, F.~Zhang, C.-L. Chang, M.~G. Yong, J.~Lee, W.-T. Chang, W.~Hua, M.~Georg, and M.~Grundmann, ``Mediapipe: A framework for building perception pipelines,'' {\em Google Research}, 2020.

\bibitem{yunet}
W.~Wu, H.~Peng, and S.~Yu, ``Yunet: A tiny millisecond-level face detector,'' {\em Department of Computer Science and Engineering, Southern University of Science and Technology, Shenzhen 518055, China}, vol.~20, no.~5, pp.~656--665, 2023.

\bibitem{yolov8}
D.~Reis, J.~Kupec, J.~Hong, and A.~Daoudi, ``Real-time flying object detection with yolov8,'' {\em Georgia Institute of Technology}, 2023.

\bibitem{deepface}
S.~I. Serengil and A.~Ozpinar, ``Lightface: A hybrid deep face recognition framework,'' in {\em 2020 Innovations in Intelligent Systems and Applications Conference (ASYU)}, (Istanbul, Turkey), pp.~1--5, 2020.

\bibitem{taigman2014deepface}
Y.~Taigman, M.~Yang, M.~Ranzato, and L.~Wolf, ``Deepface: Closing the gap to human-level performance in face verification,'' in {\em Proc. IEEE Comput. Soc. Conf. Comput. Vis. Pattern Recognit.}, pp.~1701--1708, 2014.

\bibitem{facenet}
F.~Schroff, D.~Kalenichenko, and J.~Philbin, ``Facenet: A unified embedding for face recognition and clustering,'' {\em arXiv preprint arXiv:1503.03832}, 2015.

\bibitem{vgg}
O.~M. Parkhi, A.~Vedaldi, and A.~Zisserman, ``Deep face recognition,'' {\em Visual Geometry Group, Department of Engineering Science, University of Oxford}, 2015.

\bibitem{Huang2012a}
G.~B. Huang, M.~Mattar, H.~Lee, and E.~Learned-Miller, ``Learning to align from scratch,'' in {\em NIPS}, 2012.

\bibitem{deng2018arcface}
J.~Deng, J.~Guo, J.~Yang, N.~Xue, I.~Kotsia, and S.~Zafeiriou, ``Arcface: Additive angular margin loss for deep face recognition,'' {\em arXiv preprint arXiv:1801.07698}, 2015.

\bibitem{sface}
Y.~Zhong, W.~Deng, J.~Hu, D.~Zhao, X.~Li, and D.~Wen, ``Sface: Sigmoid-constrained hypersphere loss for robust face recognition,'' 2021.

\bibitem{tracker}
J.~Nascimento, A.~Abrantes, and J.~Marques, ``An algorithm for centroid-based tracking of moving objects,'' in {\em 1999 IEEE International Conference on Acoustics, Speech, and Signal Processing. Proceedings. ICASSP99 (Cat. No.99CH36258)}, vol.~6, pp.~3305--3308 vol.6, 1999.

\bibitem{alignement}
M.~L. Cañellas, C.~Álvarez Casado, L.~Nguyen, and M.~B. López, ``Improving depression estimation from facial videos with face alignment, training optimization and scheduling,'' 2022.

\end{thebibliography}
\end{document}